\def\year{2022}\relax
\documentclass[letterpaper]{article} 
\usepackage{aaai22}  
\usepackage{times}  
\usepackage{helvet}  
\usepackage{courier}  
\usepackage[hyphens]{url}  
\usepackage{graphicx} 
\usepackage{natbib}  
\usepackage{caption} 
\DeclareCaptionStyle{ruled}{labelfont=normalfont,labelsep=colon,strut=off} 
\usepackage{algorithm}
\usepackage{algorithmic}

\usepackage{booktabs}

\usepackage{newfloat}
\usepackage{listings}
\floatstyle{ruled}
\newfloat{listing}{tb}{lst}{}
\floatname{listing}{Listing}
\title{MME-CRS: Multi-Metric Evaluation Based on Correlation Re-Scaling for Evaluating Open-Domain Dialogue}
\author {
    Pengfei Zhang\textsuperscript{\rm 1},
    Xiaohui Hu\textsuperscript{\rm 1},
    Kaidong Yu\textsuperscript{\rm 1},
    Jian Wang\textsuperscript{\rm 1},
    Song Han\textsuperscript{\rm 2},
    Cao Liu\textsuperscript{\rm 1*},
    Chunyang Yuan\textsuperscript{\rm 1}
}
\affiliations {
    \textsuperscript{\rm 1}Meituan\\
    \textsuperscript{\rm 2}Beijing University of Posts and Telecommunications\\
    \{zhangpengfei36, huxiaohui08, yukaidong, wangjian79, liucao, yuanchunyang\}@meituan.com,
    hsong@bupt.edu.cn
}

\usepackage{bibentry}

\begin{document}

\maketitle

\begin{abstract}
Automatic open-domain dialogue evaluation is a crucial component of dialogue systems. Recently, learning-based evaluation metrics have achieved state-of-the-art performance in open-domain dialogue evaluation. However, these metrics, which only focus on a few qualities, are hard to evaluate dialogue comprehensively. Furthermore, these metrics lack an effective score composition approach for diverse evaluation qualities.
To address the above problems, we propose a Multi-Metric Evaluation based on Correlation Re-Scaling (MME-CRS) for evaluating open-domain dialogue. Firstly, we build an evaluation metric composed of 5 groups of parallel sub-metrics called Multi-Metric Evaluation (MME) to evaluate the quality of dialogue comprehensively. Furthermore, we propose a novel score composition method called Correlation Re-Scaling (CRS) to model the relationship between sub-metrics and diverse qualities. Our approach MME-CRS \textbf{ranks first} on the final test data of DSTC10 track5  sub-task1 ``Automatic Open-domain Dialogue Evaluation Challenge" with a large margin, which proved the effectiveness of our proposed approach.
\end{abstract}

\section{Introduction}
\noindent Automatic evaluation is a crucial component for the development of open-domain dialogue systems \citep{danescu2011chameleons,yao2017towards}. The goal of dialogue evaluation is to produce evaluation scores that correlate well to human judgments (scores) over multiple dialogue qualities, i.e., fluency, relevancy, and specificity \citep{zhang2018learning,weston2018retrieve}.
In Dialogue System Technology Challenge 10 (DSTC10) \footnote{\url{https://dstc10.dstc.community/home}}, the track5 sub-task1 ``Automatic Open-domain Dialogue Evaluation" \citep{chen2021automatic}  proposes such a challenge in which all participants need to seek effective automatic dialogue evaluation metrics on 14 datasets (37 evaluation qualities in total) during the development phase. A single overall score must be submitted for each dialogue in test datasets during the final evaluation phase.

\begin{figure*}[t]
\centering
\includegraphics[width=2.1\columnwidth]{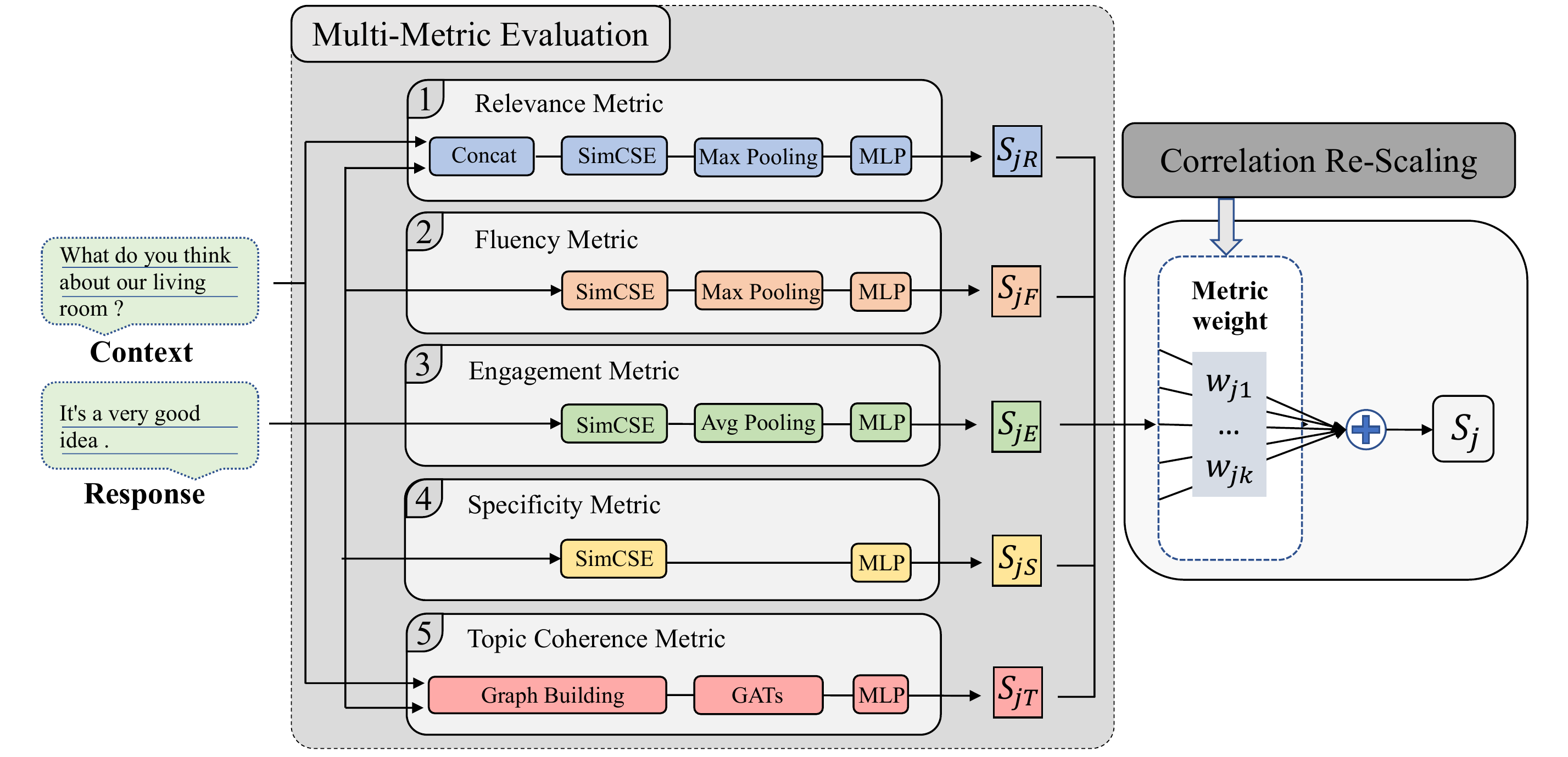} 
\caption{The architecture of the proposed evaluation approach. The ``metric weight" is pre-computed using our CRS method. The $S_j$ is the final composition score of evaluation quality $q_j$ for the context-response pair in Figure 1.}
\label{fig1}
\end{figure*}

Nowadays, word overlap-based metrics and embedding-based metrics are standard automatic evaluation metrics. Word overlap-based metrics, which measure the overlapping words between reference and candidate responses, have been used to evaluate the dialogue responses \citep{papineni2002bleu,banerjee2005meteor,sordoni2015neural}.
Embedding-based metrics measure the evaluation quality of a response by calculating the semantic similarity between the model response and corresponding reference, such as Greedy Matching \citep{rus2012optimal}, Embedding Averaging \citep{wieting2015towards}, and BERTScore \citep{zhang2019bertscore}. 

However, the aforementioned metrics heavily rely on the given references, and it has been shown that these metrics are ineffective due to the one-to-many nature of dialogue \citep{zhao2017learning}. 
Recently, learning-based metrics, which aim to predict the scores of various qualities of response, have a better correlation with human judgment \citep{tao2018ruber,ghazarian2019better,lan2020pone}.
For example, USL-H \citep{phy2020deconstruct} designs BERT-based \citep{devlin2019bert} classifiers for three groups of evaluation qualities: understandability \citep{nubel1997end}, sensibleness \citep{adiwardana2020towards}, and likability. USL-H also applies a simple weighted sum to integrate the scores of each evaluation quality. Therefore, USL-H achieves good correlations with human judgment.

Nevertheless, these metrics have some important issues in dealing with dialogue evaluation tasks.
First, most evaluation models, which only focus on a few evaluation aspects, are difficult to fully measure the quality of the open-domain dialogue. For example, USL-H ignores some important qualities like topic transition dynamics \citep{huang2020grade} and user engagement in dialogue \citep{ghazarian2020predictive}. 
Second, these metrics lack an effective score composition approach to integrate scores generated for each evaluation quality.

To address the above issues, we propose a Multi-Metric Evaluation based on Correlation Re-Scaling (MME-CRS) for evaluating open-domain dialogue as follows. 
Firstly, to evaluate the dialogue quality more comprehensively, we design 5 groups of sub-metrics for sub-task1 instead of three groups of metrics designed by USL-H. Second, we propose a novel score composition method called Correlation  Re-Scaling (CRS) to composite metric scores. Our proposed approach ranks first and achieves an average Spearman correlation score of 31.04\% on the test dataset, which is 1.11\% higher than the second. 

In particular, we summarize our contributions in this paper as follows:

\begin{itemize}
    \item We design an evaluation metric composed of 5 groups of sub-metrics to better evaluate the comprehensive quality of open-domain dialogue.
    \item We propose a novel score composition method CRS to integrate sub-metric scores more effectively. The weight distribution generated by CRS generalizes well on unseen test data.
    \item Our proposed metric MME-CRS ranks first on the ``Automatic Open-domain Dialogue Evaluation" of DSTC10 track5 task1 with a large margin, which proves the superiority of our designed metrics and CRS method.
\end{itemize}

\section{Methodology}

Figure 1 shows the architecture of our proposed metric MME-CRS. In this section, we will first introduce 5 groups of sub-metrics in detail. Then score composition approach CRS is discussed to integrate sub-metric scores for diverse qualities.

\subsection{Automatic Evaluation Metrics}
The evaluation quality contains various aspects, such as fluency, relevancy, specificity, and user engagement. For example, sub-task1 of DSTC10 track5 contains 14 development datasets, of which 37 different qualities are included in the total. What’s more, the evaluation of each aspect usually relies on several metrics, and the weight distribution over sub-metric varies from aspect to aspect. To better measure each evaluation aspect of dialogue, we design 5 groups of fundamental sub-metrics as follows.

\textbf{Fluency Metric (FM)} quantifies whether or not a response is fluency or understandable. A fluency utterance does not have to be grammatically correct because an open-domain response is usually the central part of a complete sentence. The auxiliary verb or stop words may be missing. 

We use this characteristic to build a training set of fluent and non-fluent responses. First, we randomly determine if a response $r$ is fluent. If it is, we assign response $r$ with label one and randomly apply one of the following rules: (i) no modification, (ii) delete each stopword with a probability of 0.5. Otherwise, we label response $r$ with zero and apply one of the following rules following \citet{sinha2020learning} for negative sampling: (i) word reorder (shuffle the order of all words), (ii) word drop (randomly drop x\% words), or (iii) words repeat (randomly select span(s) of words and randomly repeat them). 

For a response $r$ with $(w_1,w_2,...w_n)$ words,
we fine-tune SimCSE \citep{gao2021simcse} to embed each word in $r$ and apply Max-Pooling to get the utterance embedding. Then a Softmax layer is used to obtain the probability, and we use it as the fluency score $S_{F}$.

\textbf{Relevance Metric (RM)} measures
coarse-grained relevance between context and response.
We fine-tune another SimCSE model based on the next utterance prediction task to predict whether a context-response pair is relevant or not. Similar to the fluency metric, we first randomly determine a context-response pair from the Daily Dialog dataset \citep{li2017dailydialog} is valid or not. For the valid case, we randomly apply one of the following changes to the response: (i) no modification, (ii) remove stop words. 

\citet{lan2020pone} observes that most random sampled negative responses are low-quality, and the decision boundary learned is far from the actual decision boundary, which hurts the performance. Hence, for the invalid case, we propose a simple but effective negative response sampling method. First, we randomly choose ten responses from the response pool and compute the Word2Vec similarity \citep{mikolov2013efficient} between reference and candidate responses. Then we sort candidate responses based on their similarity score and choose the middle one as a negative response. 

To fine-tune the SimCSE model, we first concatenate a context-response pair to a single sentence. Then we compute the score $S_{R}$ using the same approach as the fluency metric.

\textbf{Topic Coherence Metric (TCM)} measures fine-grained topic transition dynamics of dialogue flows. \citet{huang2020grade} demonstrates the effectiveness of incorporation graph information into dialogue evaluation. Following \citet{huang2020grade}, topic-level dialogue graphs are firstly constructed based on ConceptNet \citep{speer2017conceptnet}. The topic transition dynamics over topic-level dialogue graphs are modeled applying a graph neural network. Then the topic-level graph representation is fed into an MLP layer to predict topic coherence score $S_T$. \citet{huang2020grade} also embeds the context-response pair and jointly predict coherence score together with topic-level graph representation. The former embedding is ignored in this part to focus on the topic coherence metric.

\textbf{Engagement Metric (EM)} measures whether the user is willing to participate in the dialogue.
We build a training set based on the human engagement scores. User engagement score usually ranges from 0 to 5, and the user’s enthusiasm is proportional to the engagement score. \citet{ghazarian2020predictive} propose to label response with engagement score less than two as zero, while we find that scaling the engagement score to between 0 and 1 yields more significant benefits.

We train an utterance-level engagement classifier to predict whether the user engagement is high or low. Specifically, for a response $r$ with $(w_1,w_2,...,w_n)$ words, we fine-tune SimCSE to get the contextual embedding $h_i$ for each word $w_i$. We use average-pooling here to get the embedding of the whole response. Then an MLP layer followed by a Softmax layer is added to predict the engagement score $S_{E}$.

\citet{ghazarian2020predictive} aggregates the embedding of both context and response to predict the score of user engagement, while we observe that user engagement mainly relies on the model response. The relationship between dialogue context and response should be handled by relevancy metric or topic coherence metric. 


\textbf{Specificity Metric (SM)} measures the model's ability to handle diverse words in complex open-domain talking context. 
We introduce the specificity metric here because some deep models tend to generate general or ambiguous answers. 

\citet{mehri2020usr} uses a Roberta model \citep{liu2019roberta} to compute the mask language model (MLM) task, while we use a more light SimCSE model following other proposed sub-metrics. Similar to \citep{phy2020deconstruct}, we only use the response $r$ with $(w_1,w_2,...,w_n)$ words to compute specific score. In detail, we mask each response word $w_i$ and predict negative log-likelihood (SM-LL) based on SimCSE-MLM. We also investigate negative cross-entropy (SM-NCE) and perplexity (SM-PPL) to further improve the effectiveness of specific metrics.


\subsection{Correlation Re-Scaling Method}
Instead of designing a score composition function for the overall aspect alone, we propose to compute weight distribution along designed sub-metrics for each evaluation aspect.
The evaluation of each evaluation aspect usually relies on several designed sub-metric. For example, suppose an annotator thinks a response generated by the dialogue model is specific. In that case, he probably implies that the response is also fluent and relevant to the dialogue context. However, the designed specific metric is only trained to predict the specific score for a response. Hence, to better evaluate each dialogue aspect, we propose to model the relationship between designed sub-metrics and diverse evaluation qualities. 

We propose a novel Correlation Re-Scaling (CRS) method to compute the weight distribution for each aspect. For a dialogue evaluation dataset $D_i$ with $(q_1,q_2,...,q_n)$ qualities, we first randomly sample 300 dialogues for Spearman correlation computation. 

For each dialogue $q_{ij}$ in dataset $D_i$, we compute fundamental sub-metric scores as $S_{ijk}$, where $k$ is the number of sub-metrics. If Spearman correlation score $S_{ijk}$ is less than 0, then the corresponding sub-metric is believed to have no contribution to dialogue quality $q_{ij}$; thus, $S_{ijk}$ is simply set to 0. We treat correlation score  $S_{ijk}$ as the importance of the corresponding sub-metric to quality $q_{ij}$.

We believe that important sub-metrics should be given higher weight, it is significant for score composition over multiple scores. Hence we compute the normalized weight distribution $w_{ijk}$ as follows:

\begin{equation}
    w_{ijk}=\frac{S_{ijk}^{d_{ij}}}{\sum_{k}S_{ijk}^{d_{ij}}}
\end{equation}

Where $d_{ij}$ is the power number of $S_{ijk}$, and the assigned weight to $S_{ijk}$ is $S_{ijk}^{d_{ij}-1}$. The larger $d_{ij}$ is, the more weight is given to more important sub-metrics. 
According to our experiments, the effect of the score composition method works best when $max(S_{ijk})$ is between 1/3 and 1/2. It is a simple but effective way to determine the value of $d_{ij}$. 

To further improve the generalization ability of CRS method, we calculate the average $w_{ijk}$ of 14 development datasets as follows:

\begin{equation}
    w_{jk}=\frac{1}{|D_{q_{j}}|}\sum_{i}w_{ijk}
\end{equation}

Where $|D_{q_j}|$ is the number of development datasets that have $q_j$ quality, and $w_{jk}$ is the normalized weight distribution over each sub-metric for diverse qualities.

For each test dataset $D_i$, we first compute 7 kinds of sub-metric scores. Then the composition score for each evaluation quality $q_{ij}$ can be computed as follows: 

\begin{equation}
    S_{ij}=\sum_{k}w_{jk}\cdot S_{ijk}
\end{equation}

\section{Experiments}

\subsection{Datasets}

\textbf{Dataset for Metric Training.} 
The organizers of the task1 require that the development datasets are only allowed for validating the proposed metrics, not for training the evaluation systems. 

Hence, we select the Daily Dialog dataset \citep{li2017dailydialog} for training our metrics (except for the user engagement metric). The Daily Dialog dataset, which is about day-to-day communication on daily topics, consists of 11118/1000/1000 dialogues for train/valid/test sets, respectively. As for the user engagement metric, we use 13124 responses from ConvAI datasets\footnote{\url{http://convai.io/2017/data/}} and scale the engagement score proportionally to between 0 and 1. 

\begin{table}[]
\centering
\begin{tabular}{lrrr}
\hline
\toprule
\textbf{Dataset} & \textbf{Turns} & \textbf{Qualities} & \textbf{Annos} \\ \toprule
DSTC6 & 40000 & 1 & 400000 \\ \hline
DSTC7 & 9900 & 1 & 29700 \\ \hline
Persona-Chatlog & 3316 & 9 & 29844 \\ \hline
PersonaChat-USR & 300 & 6 & 5400 \\ \hline
TopicalChat-USR & 360 & 6 & 6480 \\ \hline
FED-Turn & 375 & 9 & 3348 \\ \hline
FED-Conversation & 125 & 11 & 1364 \\ \hline
DailyDialog-Gupta & 500 & 1 & 1500 \\ \hline
DailyDialog-Zhao & 900 & 4 & 14400 \\ \hline
PersonaChat-Zhao & 900 & 1 & 3600 \\ \hline
DailyDialog-Grade & 300 & 1 & 3000 \\ \hline
Empathetic-Grade & 300 & 1 & 3000 \\ \hline
ConvAI2-Grade & 300 & 1 & 3000 \\ \hline
HUMOD & 9500 & 2 & 57000 \\ \bottomrule
\end{tabular}
\caption{Development dataset statistics of the DSTC10 track5 task1. The Turns column is the number of dialogue in the dataset, and the Annos column is the total number of human annotations.}
\end{table}

\begin{table}[]
\centering
\begin{tabular}{lrrr}
\hline
\toprule
\textbf{Dataset} & \textbf{Turns} & \textbf{Qualities} & \textbf{Annos} \\ \toprule
JSALT & 741 & 1 & 2822 \\ \hline
ESL & 1242 & 1 & 1242 \\ \hline
NCM & 2461 & 1 & 2461 \\ \hline
DSTC10-Topical & 4500 & 4 & 72000 \\ \hline
DSTC10-Persona & 5000 & 4 & 91360 \\ \bottomrule
\end{tabular}
\caption{Test dataset statistics of the DSTC10 track5 task1.}
\end{table}

\begin{table*}[]
\centering
\begin{tabular}{lrrrrrrrrrrrr}
\toprule
\textbf{Method} & \textbf{J-A} & \textbf{E-A} & \textbf{N-A} & \textbf{DT-A} & \textbf{DT-C} & \textbf{DT-G} & \textbf{DT-R} & \textbf{DP-A} & \textbf{DP-C} & \textbf{DP-G} & \textbf{DP-R} & \textbf{Avg} \\ \toprule
Deep AM-FM & 5.09 & 32.29 & 16.49 & 18.23 & 8.63 & \textbf{16.84} & 26.21 & 21.04 & 14.22 & 19.08 & 24.11 & 18.38 \\ \hline
Top 1 (\textbf{ours}) & 11.66 & 41.44 & \textbf{29.88} & \textbf{32.64} & 17.23 & 8.96 & \textbf{44.76} & \textbf{45.60} & \textbf{32.53} & \textbf{21.98} & \textbf{54.76} & \textbf{31.04} \\ \hline
Top 2 & 8.52 & 38.11 & 26.61 & 31.78 & \textbf{17.89} & 8.52 & 43.81 & 45.36 & 32.78 & 21.47 & 54.35 & 29.93 \\ \hline
Top 3 & \textbf{26.15} & \textbf{47.56} & 19.89 & 27.66 & 15.52 & 2.81 & 38.31 & 41.81 & 30.49 & 18.08 & 49.92 & 28.93 \\ \hline
Top 4 & 12.73 & 32.11 & 26.47 & 29.97 & 15.56 & 6.16 & 42.47 & 43.43 & 31.46 & 18.85 & 53.10 & 28.39 \\ \hline
Top 5 & 16.42 & 43.60 & 27.05 & 30.75 & 12.62 & 7.54 & 41.86 & 39.86 & 22.95 & 17.42 & 47.14 & 27.93 \\ \bottomrule
\end{tabular}
\caption{The Spearman correlation (\%) of baseline Deep AM-FM and top 5 teams on the test datasets. The test dataset JSALT, ESL, NCM, DSTC10-Topical, DSTC10-Persona are abbreviated as J, E, N, DT, DP respectively. And the evaluation qualities appropriateness, content, grammar, relevance are abbreviated as A, C, G, R respectively. The Avg column lists the average score of corresponding method on the 11 evaluation qualities.  \textbf{Bold} denotes the best result for the corresponding quality.}
\end{table*}

\begin{table}[]
\centering
\begin{tabular}{lcc}
\toprule
\textbf{Method} & \textbf{Submission Rank} & \textbf{Avg} \\ \toprule

Our submission 1 & 2 & 29.96 \\ \hline

Our submission 2 & 1 & \textbf{31.04} \\ \hline

Our submission 3 & 4 & 29.77 \\ \hline

Our submission 4 & 5 & 29.64 \\ \hline

Our submission 5 & 6 & 29.61 \\ \bottomrule
\end{tabular}
\caption{The average Spearman correlation (\%) of our different submissions on the test datasets. Each participant has up to 5 submissions, and we list the rank of our submissions in all submissions. \textbf{Bold} denotes the best result in our submissions.}
\end{table}

\textbf{Development dataset.} During the development phase, the evaluation metrics need to evaluate on 14 datasets (37 evaluation qualities in total). The organizers of ``Automatic Open-domain Dialogue Evaluation" identify the following datasets to test the effectiveness of the proposed evaluation metrics \footnote{The statistics information of development datasets is listed in Table 1.}: 

\begin{itemize}
    \item DSTC6 \citep{hori888end}.
    \item DSTC7 \citep{galley2019grounded}.
    \item Persona-Chatlog \citep{see2019makes}.
    \item PersonaChat-USR \citep{mehri2020usr}.
    \item TopicalChat-USR \citep{mehri2020usr}.
    \item FED-Turn \citep{mehri2020unsupervised}.
    \item FED-Conversation \citep{mehri2020unsupervised}.
    \item DailyDialog \citep{gupta2019investigating}.
    \item DailyDialog \citep{zhao2020designing}.
    \item PersonaChat \citep{zhao2020designing}.
    \item DailyDialog \citep{huang2020grade}.
    \item Empathetic \citep{huang2020grade}.
    \item ConvAI2 \citep{huang2020grade}.
    \item HUMOD \citep{merdivan2020human}.
\end{itemize}

\textbf{Test dataset.} During the final test phase, 5 datasets (including 11 evaluation qualities in total) are introduced by task1 organizers to fully evaluate the proposed metrics. The dataset statistics are listed in Table 2.









\subsection{Evaluation Criteria}
The Spearman correlation is used in the ``Automatic Open-domain Dialogue Evaluation Challenge" of DSTC10. The Spearman correlations between the submitted scores and corresponding human scores will be computed per evaluation category per dataset. The submissions from different participants will be ranked by the average correlation scores across all the datasets' evaluation qualities. 

\subsection{Training Details}
We use a pre-trained SimCSE model \citep{gao2021simcse} to fine-tune proposed metrics except for the topic coherence metric, and the model weights of different metrics are not shared. All the metrics based on the SimCSE model are trained with an Adam optimizer \citep{kingma2015adam} with a learning rate of 1e-5. We train these metrics on the Daily Dialog dataset \citep{li2017dailydialog} and choose models that have the lowest loss on the Daily Dialog evaluation data. We also test other pre-trained models, such as BERT \citep{devlin2019bert} or Roberta \citep{liu2019roberta}, but no performance improvement is observed.

Similar to \citet{huang2020grade}, the layer of graph attention networks (GATs) \citep{velivckovic2018graph} is 3, and the number of heads is set to 4, but we remove the contextualized encoding of context-response pair for model simplification. The training of the model is consistent with \citet{huang2020grade} except that we modify the learning rate to 1e-5.

\subsection{Overall Comparisons}
\textbf{Comparison Setting.} In this part, we will compare 1) Deep AM-FM \citep{zhang2021deep} (the baseline of task1) and top 5 teams; 2) our different submissions on the test datasets. Table 3 lists the Deep AM-FM and top 5 with the highest average Spearman correlation on the test datasets. In Table 4, we list the comparison results of our 5 submissions. The baseline method Deep AM-FM and our submissions are introduced as follows:

\begin{itemize}
    \item Deep AM-FM. A DNN-based automatic metric that measures the evaluation quality of dialogue generation along two dimensions: 1)  Adequacy Metric: The semantic similarity between dialogue context and response; 2) Fluency Metric: The syntactic quality of the sentence construction.
    \item Top1-5. Top1-5 refer to the top 5 teams with the highest average Spearman correlation on test datasets of task1. And the top1 is our team.
    \item Our submission 1-3. The metric scores are integrated using our proposed CRS score composition method. To simplify the computation of the CRS method and improve the generalization ability of our metric, we simply set $d_{ij}$ to 1, 2, 3, respectively in our submission 1-3.
    \item Our submission 4-5. We compute a specific $d_{ij}$ for each dialogue quality in each dataset. In submission 4, a SimCSE pre-trained model is used, while in submission 5, a BERT pre-trained model is used instead, avoiding that the SimCSE model does not work well on the test set.
\end{itemize}

\begin{table}[]
\centering
\begin{tabular}{lc}
\toprule
\textbf{Method} &  \textbf{Avg} \\ \toprule
MME-CRS (\textbf{ours})& \textbf{31.04} \\ \hline
w/o FM& 30.34 \\ \hline
w/o RM&  29.48 \\ \hline
w/o TCM&  27.78 \\ \hline
w/o EM&  30.05 \\ \hline
w/o SM& 30.96 \\ \hline
w/o RM+TCM&  11.07 \\ \bottomrule
\end{tabular}
\caption{MME-CRS ablation study. Our submission 2 is chosen in this ablation as the first row. FM, RM, TCM, EM, SM represent fluency metric, relevance metric, topic coherence metric, user engagement metric and specific metric respectively. ``w/o FM" means that fluency metric will not be used in the score composition. \textbf{Bold} denotes the best result in the ablation.}
\end{table}

\begin{table*}[]
\centering
\begin{tabular}{lrrrrrrrrrrrr}
\toprule
\textbf{Method}&\textbf{J-A}&\textbf{E-A}&\textbf{N-A}&\textbf{DT-A}&\textbf{DT-C}&\textbf{DT-G}&\textbf{DT-R}&\textbf{DP-A}&\textbf{DP-C}&\textbf{DP-G}&\textbf{DP-R}&\textbf{Avg} \\ \toprule
MME-CRS&\textbf{11.66}&\textbf{41.44}&\textbf{29.88}&\textbf{32.64}&\textbf{17.23}&\textbf{8.96}&\textbf{44.76}&\textbf{45.60}&\textbf{32.53}&\textbf{21.98}&\textbf{54.76}&\textbf{31.04} \\ \hline

MME-Avg&8.23 & 37.19 & 28.88 & 30.64 & 13.42 & 6.67 & 41.60 & 42.83 & 26.59 & 19.83 & 47.21&27.55 \\ \bottomrule
\end{tabular}
\caption{The comparison results of MME-CRS and MME-Avg. MME-Avg assigns equal weights to designed metrics, thus the composition score are simply the average of metric scores.}
\end{table*}

\textbf{Comparison Results.}
We compare our approach with Deep AM-FM and the top 5 teams in Table 3. The performance comparison of our different submissions is shown in Table 4.
These results support the following statements:

\begin{itemize}
    \item Our MME-CRS achieves the highest average Spearman correlation (1.11\% higher than the second) on five test datasets, which demonstrates the effectiveness of our proposed metric MME-CRS.
    \item Our method ranks first in 6 out of 11 dialogue evaluation qualities, demonstrating that our proposed evaluation metrics have a higher correlation with human judgments than baseline and other teams.
    \item Our submissions 1-3, which fix $d_{ij}$ to a constant number, perform better than submissions 4-5, which indicate a constant $d$ generalizes better when migrating to the test datasets. Furthermore, setting $d_{ij}$ to 2 achieves the best performance on test datasets.
\end{itemize}

\subsection{Ablation Study}
\textbf{Comparison Setting.}
In the final evaluation period, participants must submit a single score for each dialogue, and the organizers will compute the correlation between human scores of each quality with submitted scores. In Table 5, we remove fluency metric (FM), relevance metric (RM), topic coherence metric (TCM), engagement metric (EM), and specific metric (SM), respectively, to explore the importance of different metrics in the submitted scores. We take the SM composed of three metrics as a whole part to explore the influence of the SM. Considering that RM and TCM both rely on dialogue context, we also remove them together in our experiments.

\textbf{Comparison Results.}
The comparison results of the ablation experiment are shown in Table 5. These results support the following statements:

\begin{itemize}
    \item TCM, RM, and EM contribute most to the performance. When we delete them from score composition, the final average Spearman correlation will drop 3.26\%, 1.56\%, and 1.01\%, respectively. 
    \item Coarse-grained RM and fine-grained TCM are a beneficial complement to each other. If we ignore one of them, the performance will drop slightly. However, if both of them are ignored, the average correlation will drastically drop to 11.07\%.
    \item The improvement of SM can be ignored on the test datasets. We observe that many responses in test datasets tend to be very specific but are not relevant to the dialogue context. 
    We infer that these models used to generate the response are overfitted on the test dataset.
\end{itemize}

\subsection{The Effectiveness of CRS Method}
\textbf{Comparison Setting.}
Score composition is a significant component of open-domain dialogue because the full evaluation of dialogue usually depends on many aspects. In this part, we will compare the performance of MME-Avg, which simply averages the scores from different metrics, and our MME-CRS method.

\textbf{Comparison Results.}
The comparison result is listed in Table 6, and the following statements can be drawn from the results.

\begin{itemize}
    \item The average Spearman correlation score is significantly superior to that of MME-Avg (3.49\% higher), indicating that our proposed CRS method can effectively integrate different scores to comprehensively measure the quality of dialogue. 
    \item The correlation of MME-CRS is higher on all evaluation qualities, demonstrating that each evaluation quality can benefit from the score composition method CRS.
\end{itemize}

\subsection{}

\section{Conclusion}
In this paper, we propose an open-domain dialogue evaluation approach composed of 5 groups of metrics to fully measure the quality of model response. Further, we propose a novel metric composition method called CRS. CRS models the relationship between metrics and evaluation qualities to comprehensively integrate sub-metric scores for dialogue evaluation. Experimental results on test datasets show that our proposed MME-CRS achieves the best performance, showing that our metric correlates better with human judgments. Compared with baseline and other teams, our approach obtains superior performance and ranks 1st in the final evaluation.


\bibliography{aaai22}

\begin{thebibliography}{38}
\providecommand{\natexlab}[1]{#1}

\bibitem[{Adiwardana et~al.(2020)Adiwardana, Luong, So, Hall, Fiedel,
  Thoppilan, Yang, Kulshreshtha, Nemade, Lu et~al.}]{adiwardana2020towards}
Adiwardana, D.; Luong, M.-T.; So, D.~R.; Hall, J.; Fiedel, N.; Thoppilan, R.;
  Yang, Z.; Kulshreshtha, A.; Nemade, G.; Lu, Y.; et~al. 2020.
\newblock Towards a human-like open-domain chatbot.
\newblock \emph{arXiv preprint arXiv:2001.09977}.

\bibitem[{Banerjee and Lavie(2005)}]{banerjee2005meteor}
Banerjee, S.; and Lavie, A. 2005.
\newblock METEOR: An automatic metric for MT evaluation with improved
  correlation with human judgments.
\newblock In \emph{Proceedings of the acl workshop on intrinsic and extrinsic
  evaluation measures for machine translation and/or summarization}, 65--72.

\bibitem[{Chen et~al.(2021)Chen, Sadoc, D'Haro, Banchs, and
  Rudnicky}]{chen2021automatic}
Chen, Z.; Sadoc, J.; D'Haro, L.~F.; Banchs, R.; and Rudnicky, A. 2021.
\newblock Automatic evaluation and moderation of open-domain dialogue systems.
\newblock \emph{arXiv preprint arXiv:2111.02110}.

\bibitem[{Danescu-Niculescu-Mizil and Lee(2011)}]{danescu2011chameleons}
Danescu-Niculescu-Mizil, C.; and Lee, L. 2011.
\newblock Chameleons in imagined conversations: a new approach to understanding
  coordination of linguistic style in dialogs.
\newblock In \emph{Proceedings of the 2nd Workshop on Cognitive Modeling and
  Computational Linguistics}, 76--87.

\bibitem[{Devlin et~al.(2019)Devlin, Chang, Lee, and
  Toutanova}]{devlin2019bert}
Devlin, J.; Chang, M.-W.; Lee, K.; and Toutanova, K. 2019.
\newblock Bert: Pre-training of deep bidirectional transformers for language
  understanding.
\newblock In \emph{Proceedings of the 2019 Conference of the North American
  Chapter of the Association for Computational Linguistics}, 4171--4186.

\bibitem[{Galley et~al.(2019)Galley, Brockett, Gao, Gao, and
  Dolan}]{galley2019grounded}
Galley, M.; Brockett, C.; Gao, X.; Gao, J.; and Dolan, B. 2019.
\newblock Grounded response generation task at dstc7.
\newblock In \emph{AAAI Dialog System Technology Challenges Workshop}.

\bibitem[{Gao, Yao, and Chen(2021)}]{gao2021simcse}
Gao, T.; Yao, X.; and Chen, D. 2021.
\newblock SimCSE: Simple Contrastive Learning of Sentence Embeddings.
\newblock \emph{arXiv preprint arXiv:2104.08821}.

\bibitem[{Ghazarian et~al.(2019)Ghazarian, Wei, Galstyan, and
  Peng}]{ghazarian2019better}
Ghazarian, S.; Wei, J.; Galstyan, A.; and Peng, N. 2019.
\newblock Better automatic evaluation of open-domain dialogue systems with
  contextualized embeddings.
\newblock In \emph{Proceedings of the Workshop on Methods for Optimizing and
  Evaluating Neural Language Generation}, 82--89.

\bibitem[{Ghazarian et~al.(2020)Ghazarian, Weischedel, Galstyan, and
  Peng}]{ghazarian2020predictive}
Ghazarian, S.; Weischedel, R.; Galstyan, A.; and Peng, N. 2020.
\newblock Predictive engagement: An efficient metric for automatic evaluation
  of open-domain dialogue systems.
\newblock In \emph{Proceedings of the AAAI Conference on Artificial
  Intelligence}, volume~34, 7789--7796.

\bibitem[{Gupta et~al.(2019)Gupta, Mehri, Zhao, Pavel, Eskenazi, and
  Bigham}]{gupta2019investigating}
Gupta, P.; Mehri, S.; Zhao, T.; Pavel, A.; Eskenazi, M.; and Bigham, J.~P.
  2019.
\newblock Investigating evaluation of open-domain dialogue systems with human
  generated multiple references.
\newblock In \emph{Proceedings of the 20th Annual SIGdial Meeting on Discourse
  and Dialogue}, 379--391.

\bibitem[{Hori and Hori(2017)}]{hori888end}
Hori, C.; and Hori, T. 2017.
\newblock End-to-end Conversation Modeling Track in DSTC6.
\newblock \emph{dialog}, 888(107,506): 2--000.

\bibitem[{Huang et~al.(2020)Huang, Ye, Qin, Lin, and Liang}]{huang2020grade}
Huang, L.; Ye, Z.; Qin, J.; Lin, L.; and Liang, X. 2020.
\newblock Grade: Automatic graph-enhanced coherence metric for evaluating
  open-domain dialogue systems.
\newblock In \emph{Proceedings of the 2020 Conference on Empirical Methods in
  Natural Language Processing (EMNLP)}, 9230--9240.

\bibitem[{Kingma and Ba(2015)}]{kingma2015adam}
Kingma, D.~P.; and Ba, J. 2015.
\newblock Adam: A method for stochastic optimization.
\newblock In \emph{ICLR (Poster)}.

\bibitem[{Lan et~al.(2020)Lan, Mao, Wei, Gao, and Huang}]{lan2020pone}
Lan, T.; Mao, X.-L.; Wei, W.; Gao, X.; and Huang, H. 2020.
\newblock Pone: A novel automatic evaluation metric for open-domain generative
  dialogue systems.
\newblock \emph{ACM Transactions on Information Systems (TOIS)}, 39(1): 1--37.

\bibitem[{Li et~al.(2017)Li, Su, Shen, Li, Cao, and Niu}]{li2017dailydialog}
Li, Y.; Su, H.; Shen, X.; Li, W.; Cao, Z.; and Niu, S. 2017.
\newblock Dailydialog: A manually labelled multi-turn dialogue dataset.
\newblock In \emph{Proceedings of the Eighth International Joint Conference on
  Natural Language Processing (Volume 1: Long Papers)}, 986--995.

\bibitem[{Liu et~al.(2019)Liu, Ott, Goyal, Du, Joshi, Chen, Levy, Lewis,
  Zettlemoyer, and Stoyanov}]{liu2019roberta}
Liu, Y.; Ott, M.; Goyal, N.; Du, J.; Joshi, M.; Chen, D.; Levy, O.; Lewis, M.;
  Zettlemoyer, L.; and Stoyanov, V. 2019.
\newblock Roberta: A robustly optimized bert pretraining approach.
\newblock \emph{arXiv preprint arXiv:1907.11692}.

\bibitem[{Mehri and Eskenazi(2020{\natexlab{a}})}]{mehri2020unsupervised}
Mehri, S.; and Eskenazi, M. 2020{\natexlab{a}}.
\newblock Unsupervised evaluation of interactive dialog with dialogpt.
\newblock In \emph{Proceedings of the 21th Annual Meeting of the Special
  Interest Group on Discourse and Dialogue}, 225--235.

\bibitem[{Mehri and Eskenazi(2020{\natexlab{b}})}]{mehri2020usr}
Mehri, S.; and Eskenazi, M. 2020{\natexlab{b}}.
\newblock Usr: An unsupervised and reference free evaluation metric for dialog
  generation.
\newblock In \emph{Proceedings of the 58th Annual Meeting of the Association
  for Computational Linguistics}, 681--707.

\bibitem[{Merdivan et~al.(2020)Merdivan, Singh, Hanke, Kropf, Holzinger, and
  Geist}]{merdivan2020human}
Merdivan, E.; Singh, D.; Hanke, S.; Kropf, J.; Holzinger, A.; and Geist, M.
  2020.
\newblock Human annotated dialogues dataset for natural conversational agents.
\newblock \emph{Applied Sciences}, 10(3): 762.

\bibitem[{Mikolov et~al.(2013)Mikolov, Chen, Corrado, and
  Dean}]{mikolov2013efficient}
Mikolov, T.; Chen, K.; Corrado, G.; and Dean, J. 2013.
\newblock Efficient estimation of word representations in vector space.
\newblock \emph{arXiv preprint arXiv:1301.3781}.

\bibitem[{N{\"u}bel(1997)}]{nubel1997end}
N{\"u}bel, R. 1997.
\newblock End-to-End evaluation in VERBMOBIL I.
\newblock \emph{Proceedings of MT Summit VI}, 232--239.

\bibitem[{Papineni et~al.(2002)Papineni, Roukos, Ward, and
  Zhu}]{papineni2002bleu}
Papineni, K.; Roukos, S.; Ward, T.; and Zhu, W.-J. 2002.
\newblock Bleu: A method for automatic evaluation of machine translation.
\newblock In \emph{Proceedings of the 40th annual meeting of the Association
  for Computational Linguistics}, 311--318.

\bibitem[{Phy, Zhao, and Aizawa(2020)}]{phy2020deconstruct}
Phy, V.; Zhao, Y.; and Aizawa, A. 2020.
\newblock Deconstruct to reconstruct a configurable evaluation metric for
  open-domain dialogue systems.
\newblock In \emph{Proceedings of the 28th International Conference on
  Computational Linguistics}, 4164--4178.

\bibitem[{Rus and Lintean(2012)}]{rus2012optimal}
Rus, V.; and Lintean, M. 2012.
\newblock An optimal assessment of natural language student input using
  word-to-word similarity metrics.
\newblock In \emph{International Conference on Intelligent Tutoring Systems},
  675--676. Springer.

\bibitem[{See et~al.(2019)See, Roller, Kiela, and Weston}]{see2019makes}
See, A.; Roller, S.; Kiela, D.; and Weston, J. 2019.
\newblock What makes a good conversation? How controllable attributes affect
  human judgments.
\newblock In \emph{Proceedings of the 2019 Conference of the North American
  Chapter of the Association for Computational Linguistics: Human Language
  Technologies, Volume 1 (Long and Short Papers)}, 1702--1723.

\bibitem[{Sinha et~al.(2020)Sinha, Parthasarathi, Wang, Lowe, Hamilton, and
  Pineau}]{sinha2020learning}
Sinha, K.; Parthasarathi, P.; Wang, J.; Lowe, R.; Hamilton, W.~L.; and Pineau,
  J. 2020.
\newblock Learning an unreferenced metric for online dialogue evaluation.
\newblock In \emph{Proceedings of the 58th Annual Meeting of the Association
  for Computational Linguistics}, 2430--2441.

\bibitem[{Sordoni et~al.(2015)Sordoni, Galley, Auli, Brockett, Ji, Mitchell,
  Nie, Gao, and Dolan}]{sordoni2015neural}
Sordoni, A.; Galley, M.; Auli, M.; Brockett, C.; Ji, Y.; Mitchell, M.; Nie,
  J.-Y.; Gao, J.; and Dolan, W.~B. 2015.
\newblock A Neural Network Approach to Context-Sensitive Generation of
  Conversational Responses.
\newblock In \emph{Proceedings of the 2015 Conference of the North American
  Chapter of the Association for Computational Linguistics: Human Language
  Technologies}, 196--205.

\bibitem[{Speer, Chin, and Havasi(2017)}]{speer2017conceptnet}
Speer, R.; Chin, J.; and Havasi, C. 2017.
\newblock Conceptnet 5.5: An open multilingual graph of general knowledge.
\newblock In \emph{Thirty-first AAAI conference on artificial intelligence}.

\bibitem[{Tao et~al.(2018)Tao, Mou, Zhao, and Yan}]{tao2018ruber}
Tao, C.; Mou, L.; Zhao, D.; and Yan, R. 2018.
\newblock Ruber: An unsupervised method for automatic evaluation of open-domain
  dialog systems.
\newblock In \emph{Thirty-Second AAAI Conference on Artificial Intelligence}.

\bibitem[{Veli{\v{c}}kovi{\'c} et~al.(2018)Veli{\v{c}}kovi{\'c}, Cucurull,
  Casanova, Romero, Li{\`o}, and Bengio}]{velivckovic2018graph}
Veli{\v{c}}kovi{\'c}, P.; Cucurull, G.; Casanova, A.; Romero, A.; Li{\`o}, P.;
  and Bengio, Y. 2018.
\newblock Graph attention networks.
\newblock In \emph{International Conference on Learning Representations}.

\bibitem[{Weston, Dinan, and Miller(2018)}]{weston2018retrieve}
Weston, J.; Dinan, E.; and Miller, A. 2018.
\newblock Retrieve and refine: Improved sequence generation models for
  dialogue.
\newblock In \emph{Proceedings of the 2018 EMNLP Workshop SCAI: The 2nd
  International Workshop on Search-Oriented Conversational AI}, 87--92.

\bibitem[{Wieting et~al.(2016)Wieting, Bansal, Gimpel, and
  Livescu}]{wieting2015towards}
Wieting, J.; Bansal, M.; Gimpel, K.; and Livescu, K. 2016.
\newblock Towards universal paraphrastic sentence embeddings.
\newblock In \emph{4th International Conference on Learning Representations}.

\bibitem[{Yao et~al.(2017)Yao, Zhang, Feng, Zhao, and Yan}]{yao2017towards}
Yao, L.; Zhang, Y.; Feng, Y.; Zhao, D.; and Yan, R. 2017.
\newblock Towards implicit content-introducing for generative short-text
  conversation systems.
\newblock In \emph{Proceedings of the 2017 conference on empirical methods in
  natural language processing}, 2190--2199.

\bibitem[{Zhang et~al.(2021)Zhang, D’Haro, Banchs, Friedrichs, and
  Li}]{zhang2021deep}
Zhang, C.; D’Haro, L.~F.; Banchs, R.~E.; Friedrichs, T.; and Li, H. 2021.
\newblock Deep AM-FM: Toolkit for automatic dialogue evaluation.
\newblock In \emph{Conversational Dialogue Systems for the Next Decade},
  53--69. Springer.

\bibitem[{Zhang et~al.(2018)Zhang, Guo, Fan, Lan, Xu, and
  Cheng}]{zhang2018learning}
Zhang, R.; Guo, J.; Fan, Y.; Lan, Y.; Xu, J.; and Cheng, X. 2018.
\newblock Learning to control the specificity in neural response generation.
\newblock In \emph{Proceedings of the 56th Annual Meeting of the Association
  for Computational Linguistics (Volume 1: Long Papers)}, 1108--1117.

\bibitem[{Zhang et~al.(2019)Zhang, Kishore, Wu, Weinberger, and
  Artzi}]{zhang2019bertscore}
Zhang, T.; Kishore, V.; Wu, F.; Weinberger, K.~Q.; and Artzi, Y. 2019.
\newblock BERTScore: Evaluating text generation with BERT.
\newblock In \emph{International Conference on Learning Representations}.

\bibitem[{Zhao, Lala, and Kawahara(2020)}]{zhao2020designing}
Zhao, T.; Lala, D.; and Kawahara, T. 2020.
\newblock Designing precise and robust dialogue response evaluators.
\newblock In \emph{Proceedings of the 58th Annual Meeting of the Association
  for Computational Linguistics}, 26--33.

\bibitem[{Zhao, Zhao, and Eskenazi(2017)}]{zhao2017learning}
Zhao, T.; Zhao, R.; and Eskenazi, M. 2017.
\newblock Learning discourse-level diversity for neural dialog models using
  conditional variational autoencoders.
\newblock In \emph{Proceedings of the 55th Annual Meeting of the Association
  for Computational Linguistics (Volume 1: Long Papers)}, 654--664.

\end{thebibliography}
\end{document}